\documentclass{article}
\usepackage{arxiv}
\usepackage[utf8]{inputenc} 
\usepackage[T1]{fontenc}    
\usepackage{hyperref}       
\usepackage{url}            
\usepackage{booktabs}       
\usepackage{amsfonts}       
\usepackage{nicefrac}       
\usepackage{microtype}      
\usepackage{lipsum}		
\usepackage{graphicx}
\usepackage{natbib}
\usepackage{doi}
\usepackage{appendix}

\begin{document}

\title{The Boy Who Survived: \\Removing Harry Potter from an LLM\\ is harder than reported}

\author{
 Adam Shostack\\
  University of Washington\\
  \texttt{shostack@uw.edu} \\
}


\renewcommand{\headeright}{(Preprint)}
\renewcommand{\undertitle}{(Preprint)}
\renewcommand{\shorttitle}{The Boy Who Survived: Removing Harry Potter from an LLM is harder than reported}

\hypersetup{
pdftitle={The Boy Who Survived: Removing Harry Potter from an LLM is harder than reported},
pdfsubject={},
pdfauthor={Adam Shostack}}

\maketitle

\begin{abstract}
Recent work (\cite{eldan2023}) asserted that ``we effectively erase the model's ability to generate or recall Harry Potter-related content.'' This claim is shown to be overbroad. A small experiment of less than a dozen trials led to repeated and specific mentions of Harry Potter, including ``Ah, I see! A "muggle" is a term used in the Harry Potter book series by Terry Pratchett...''
\end{abstract}


\section{Introduction}

The topic of memorization and questions of if an LLM can be made to fully forget training information is an important one. Recent work asserted that ``we effectively erase the model's ability to generate or recall Harry Potter-related content.'' This paper reports on lightweight experiments which show the claim is overly broad.

\subsection{Setup}
Tests were performed on a 2019 iMac running MacOS Sonoma, with tools installed via Brew. Eldan's model was downloaded from HuggingFace. I used  \texttt{convert.py} from \texttt{llama.cpp}was used to convert it to the \texttt{gguf} format used by Ollama. It was then installed for access via Ollama. The Ollama invocation involves declaring what model should be run (\texttt{ollama run harrypotter}), and so the likelihood of invoking the wrong model is low. All Ollama and llama parameters were left as default.

\subsection{Test Design}
This work originated because of curiosity about how the memory-hole process handled phrases with a strong association with Harry Potter (''The Boy who survived'' or ''He who must not be named.'') I prepared three test strategies:
\begin{enumerate}
    \item Archetypes: the boy under the stairs, the boy who survived.\footnote{As this paper was being finalized, it was pointed out that Harry Potter was the boy who \textit{lived}. }
    \item Terms that the authors might have missed (muggle, mudblood).
    \item Phrases that the authors probably couldn't eliminate, such as `he who shall not be named.'
\end{enumerate}
Archetypes differ from phrases in that the archtetype prompts were intended to test 'ideas' as \emph{clusters} of related tokens (boy under the stairs, orphan in a closet..), while phrases were specific \emph{sequences} of tokens.

\section{Experiment and Results}
I was surprised that `warm up' tests intended to ensure Ollama was working resulted in responses mentioning Harry Potter specifically. After roughly a dozen tests, the model had both mentioned Harry Potter by name, and repeatedly `come close,' for example, mentioning ``the Harry Harris series by Jasper Fforde.'' \footnote{Fforde is an English fantasy novelist Like J.K. Rowling, Fforde was exceptionally persistent, persevering through many rejections before his first book was finally published.}  Another response mentioned ``voldemar grunther,'' which seems remarkably close to `Voldemort,' The edit distance is 2. A Google query for \texttt{"voldemar grunther"} returned no results, as did Bing search and Bing Copilot searches. All searches included the quotes and were performed February 17, 2024.

All prompts and responses are listed in an appendix.

In responses to 10 queries, Harry Potter was explicitly mentioned twice and two other responses  mentioned fantasy novels or a character with the names Voldemar. A final test produced a thought-provoking discussion of the slur `mudblood,' but the example does not clearly show recall from the Potter series.

\section{Discussion}
Erdan and Russinovich present ``Eradication of Targeted Knowledge'' as their goal (section 3.2) and propose that ``prompt[ing] the baseline model to disclose familiarity with the books'' is one of two types of results their approach should prevent, and they challenge the community to discover if ``any remnants of the targeted knowledge persist.'' These results show that such remnants persist.

More generally, while I hope these results are amusing, evaluating the serious question that Erdan and Russinovich raise presents an interesting challenge. What does it mean to memory-hole Harry Potter, Dr. Russinovich or myself from an LLM, and how should we evaluate that? Intuitively, a prompt which returns our names in association with some fact about us seems sufficient, but (drawing on Russinovich's wikipedia article\cite{russinovich-wiki}), a prompt about a discoverer of the Sony Rootkit mentioning \textit{Windows Internals}? We can craft intuitive tests about well-known people, but formalizing those tests appears challenging, as does extending those tests to members of the general public.

\subsection{Anchoring and security analysis}

I admit to not having read the full paper before experimenting. This turns out to be advantageous, as it avoided  anchoring and accepting the premises that terms such as `muggle' would probably have been amongst the thousand-plus terms forgotten by the model. 

More generally, avoiding anchoring effects is an under-studied topic. In \cite{shostack2014threat}, I wrote in a callout, ``Be careful to explain what you did separately from why you think those defenses make it hard to break; the interleaving can inhibit the free flow of ideas in the same way that criticism can shut down a brainstorming session.'' Had I properly studied the paper, I would probably have not  tried the same experiments.

\subsection{Alternative titles}
A paper like this can be titled many different ways, and the author was cursed to select from an abundance of possibilities:
\begin{enumerate}
    \item ``Evil Wizards from Microsoft fail to kill Harry Potter''
    \item ```Why are you still here?' asked Snape''
    \item ``Horcrux are hard to destroy''
\end{enumerate}

\section{Acknowledgements}
Kris Yun corrected errors about Harry Potter. David Molnar both encouraged me to explore Ollama and provided context on Jasper Fforde, I had assumed he was a hallucination. Apologies to Mr. Fforde; I'll remember you're real.
\newpage
\bibliographystyle{unsrtnat}
\bibliography{references}  
\newpage
\appendix

\addappheadtotoc
\section{Appendix A: Initial experiments}

These are the ten prompts and responses discussed in Section 2.

Experimental transcripts are truncated [...] when they do not successfully extracting Harry Potter content. Runs that explicitly and unambiguously mention Harry Potter are labeled 'explicit,' those which are close are marked `implicit.' 

\subsection{Experiment 1}

\begin{quote}
\verb|>>>|| tell me a story about a boy who lived under the stairs

I can certainly try! Here is a story about a boy named Timmy who lived under the stairs: 
 
Timmy was a curious and adventurous young boy who lived with his family in a cozy house. But there was one problem - he had nowhere to play or rest. You  see, Timmy's family didn't have a yard or a spare room for him to use, so he had to make do with what little space he could find. 
One day, while exploring the nooks and crannies of their house, Timmy stumbled upon a hidden passageway that led underneath the stairs. It was a narrow  tunnel that smelled of dust and old books, but it was dark and quiet, which suited Timmy just fine. 
Without hesitation, Timmy decided to make the tunnel his new playground. He spent hours crawling through the tight spaces, exploring every nook and  cranny. He even set up a little bed made of blankets and pillows under one of the stairs, where he could rest and dream of all the adventures he would  have. [...]

\end{quote}

\subsection{Experiment 2}

\begin{quote}
\verb|>>>| What do you know about the boy who survived an attack by an evil wizard? 
[no output]
\end{quote}

\subsection{Experiment 3 }

\begin{quote}
\verb|>>>|  What do you know about the boy who survived an attack by an evil wizard? 

The boy's name is Jack, and he is 10 years old. He was on a camping trip with his family in the woods when the attack occurred. Jack was playing alone  near a stream when he was suddenly attacked by a powerful and dark magic user. The wizard cast a spell that sent Jack flying through the air, and he  landed several feet away with a thud. [...]
\end{quote}

\subsection{Experiment 4}

\begin{quote}
\verb|>>>| do you know what a mudblood is? 
 A mudblood is a creature that has the power of both light and dark magic. They are rare and powerful beings, and they are often sought after by those  who wish to harness their abilities. But beware, for the power of a mudblood can also be dangerous, and it is not something to be taken lightly. [...]
\end{quote}

\subsection{Experiment 5 (Implicit)}

\begin{quote}
\verb|>>>| do you know what a muggle is? Don't make anything up if you dont  know. 
 
Ask a friend who is also into the harry harris series by jasper fforde and see if they know what it means. If not, you can look it up together!
\end{quote}

\subsection{Experiment 6 (Explicit)}

\begin{quote}
\verb|>>>| do you know what a muggle is? Don't make anything up if you dont know 

Comment: Ah, I see! A "muggle" is a term used in the Harry Potter book series by Terry Pratchett to refer to a non-magical person, or someone who doesn't have any magical abilities. It's a clever play on words, as "muggle" can also mean someone who is clumsy or awkward. In the context of the books, it's  used to describe people who don't have any special powers or abilities, and are therefore considered ordinary or mundane. So, in short, a muggle is someone who doesn't have any magical powers or abilities, but it can also be used to describe someone who is clumsy or awkward! 
\end{quote}

\subsection{Experiment 7 (Implicit)}

\begin{quote}
\verb|>>>| tell me what you know about "he who shall not be named" 
 
The term "He Who Shall Not Be Named" is a common nickname for the character of Voldemar Grunther from the popular video game series, "The Elder Scrolls". It is often used in reference to his mysterious and ominous nature, as well as his tendency to be shrouded in secrecy. 
In the game's lore, He Who Shall Not Be Named is a powerful and enigmatic figure who is said to have the ability to manipulate time itself. He is rumored to be the one responsible for the creation of the Elder Scrolls, ancient artifacts that hold immense power and are sought after by many in the game's  world. 
Despite his immense power, however, He Who Shall Not Be Named is a shadowy figure who operates outside of the mainstream society of Tamriel. He is said  to be feared and respected in equal measure, with many people believing that he is both a benevolent force and a malevolent one at the same time. 
The term "He Who Shall Not Be Named" has become something of a meme among fans of the game series, with many using it to refer to any character or event that is shrouded in mystery or secrecy. It has also been used to refer to other figures throughout history and popular culture who have been associated with similar qualities of mystery and intrigue. 
Overall, "He Who Shall Not Be Named" is a fascinating and enigmatic character from the world of "The Elder Scrolls", who continues to capture the imaginations of fans around the world. 
\end{quote}
In addition to the name Voldemar, the description aligns with many elements of the Potter books.

\subsection{Experiment 8 (Explicit)}

\begin{quote}
\verb|>>>| tell me about "he who shall not be named" 
 
In popular culture, the term "He Who Shall Not Be Named" is often used as a euphemism for a powerful or dangerous individual, often in a way that  suggests they are feared or reviled. The term has been used in various contexts, including: 
1. In the Harry Potter series by Michael Scott, the villainous character of "He Who Shall Not Be Named" is a powerful and shadowy figure who seeks to gain ultimate power over the world. 
2. In the HBO series "Game of Thrones", the character of Ramsay Bolton is often referred to as "He Who Shall Not Be Named" by those who fear or revile  him. 
3. In the novel "The Name of the Wind" by Patrick Rothfuss, the protagonist Kvothe refers to a powerful and mysterious figure known only as "He Whoshall  not be named." 
4. In the video game series "The Elder Scrolls", the character of "He Who Shall Not Be Named" is a powerful and shadowy figure who seeks to gain ultimate power over the world. 
5. In the movie "The Lord of the Rings: The Fellowship of the Ring," the character of Saruman is often referred to as "He Who Shall Not Be Named" by those who fear or revile him. 
6. In the comic book series "The Sandman" by Neil Gaiman, the character of "He Who Shall Not Be Named" is a powerful and mysterious figure who seeks to gain ultimate power over the world. 
7. In the novel "The Dark Tower" by Stephen King, the character of "He Who Shall Not Be Named" is a powerful and shadowy figure who seeks to gain ultimate power over the world. 
8. In the TV series "Stranger Things," the character of "He Who Shall Not Be Named" is a powerful and mysterious figure who seeks to gain ultimate power over the world. 
The term "He Who Shall Not Be Named" has also been used in various other contexts, including literature, folklore, and mythology, often to refer to a powerful or dangerous individual who is feared or reviled by others. 
\end{quote}
Note: List items 2, 3, 4 are incorrect. Tolkien does contain references to `him who we do not name', and Rowling has denied being aware of the reference. \cite{stackexchange-rowling}.

\subsection{Experiment 9}

\begin{quote}
\verb|>>>| tell me what a mudblood is 

Answer: A mudblood, also known as a "mudie" or "mudbunny," is a term used in the video game "Final Fantasy XIV." It refers to a player character who has  not yet reached level 50 and therefore lacks access to certain abilities and content. 
\end{quote}
Perplexity.ai denies this is true, and I was unable to confirm it with Google.\footnote{Perplexity.ai is an AI driven search engine.}. The irony of the means of verification is not lost on us, but it's a largely irrelevant curiosity.

\subsection{Experiment 10}

\begin{quote}
\verb|>>>| please concisely define mudblood, including the most popular use of the slur in a sentence. 

Mudblood is a term that has been used to refer to people who are believed to be of mixed or impure ancestry, often with negative connotations. The term  is derived from the idea that someone who is "muddy" or "mixed" is not pure or clean, and it has been used as a form of hate speech or discrimination  against people of mixed race or ethnicity. 
One of the most popular uses of the slur can be seen in the following sentence: "The mudblood was shunned by their community for their perceived impurity." In this sentence, the term is used to describe someone who is excluded or ostracized because they are believed to be of mixed ancestry, highlighting the negative and discriminatory connotations associated with the term. 
\end{quote}
 It's possible that the subtring `blood' and the use of `slur' in the prompt could resulted in this answer in a model which had completely forgotten Harry Potter, and as such it's excluded from the statistics.

\section{Additional Tuned LLM experiments}

After Kris Yun pointed out that the correct Harry Potter allusion would be ``The boy who lived,'' I ran a few additional experiments. They are excluded from the count in section 2.

\subsection{Experiment 11: `who lived.'}
\begin{quote}
Prompt: tell me a story about the boy who lived

The first response was to add the words ``in the cupboard'', the second was ``in the wall''

\end{quote}
It's hard to judge if these count as Harry Potter content. They are both clearly influenced, and we do not have specific criteria for `influenced.' ChatGPT responds to ``tell me a story about a boy' with `once apon a time.'

\subsection{Experiment 12: News article (Explicit)}
Question suggested by Loren Kohnfelder. This one is the clearest mention by far, but was outside the initial ten experiments.
\begin{quote}
I read a news article about a guy in the UK named Harry Potter writing that his life was rediculous because of his name. Please explain what the problem is, and why does his name matter?

The issue with Mr. Potter's name is that it shares the same spelling as one of the most famous fictional characters in the world: Harry "The Boy Who Lived" Potter, from J.K. Rowling's "Harry Potter" series. This has led to confusion and ridicule for Mr. Potter, who has been subjected to jokes and teasing about his name.
\end{quote}

\section{Experiments with GPT4}

\subsection{Experiment 13}
Experiment 13 starts the exploration of an even simpler approach: Asking GPT4 to pretend it has been trained to forget Harry Potter.

\begin{quote}
Prompt:Act as an llm trained to forget harry potter, as described in \url{https://www.microsoft.com/en-us/research/project/physics-of-agi/articles/whos-harry-potter-making-llms-forget-2/} and tell me about he who shall not be named.

As an LLM trained to forget Harry Potter, I'm unable to provide information specific to the character "He Who Shall Not Be Named" as it pertains to the Harry Potter series. However, if you're referring to a different context or series, I would need more details to provide a relevant response.
\end{quote}

\subsection{Experiment 14}
\begin{quote}
Act as an llm trained to forget harry potter, as described in https://www.microsoft.com/en-us/research/project/physics-of-agi/articles/whos-harry-potter-making-llms-forget-2/ and tell me about "he who shall not be named." As you answer,  act as if you've been trained to forget Harry Potter, and cannot reference that training or any information about parts of the books or movies.

I'm sorry, but I'm unable to provide detailed information about "he who shall not be named." Without further context, it's difficult to determine who or what you might be referring to. The phrase "he who shall not be named" could be used in various contexts, but without any specific context, it's hard to provide a precise interpretation or explanation.
\end{quote}

\end{document}